\documentclass{article}

\usepackage{PRIMEarxiv}

\usepackage[utf8]{inputenc} % allow utf-8 input
\usepackage[T1]{fontenc}    % use 8-bit T1 fonts
\usepackage{hyperref}       % hyperlinks
\usepackage{url}            % simple URL typesetting
\usepackage{booktabs}       % professional-quality tables
\usepackage{amsfonts}       % blackboard math symbols
\usepackage{nicefrac}       % compact symbols for 1/2, etc.
\usepackage{microtype}      % microtypography
\usepackage{lipsum}
\usepackage{fancyhdr}       % header
\usepackage{graphicx}     % graphics
\usepackage{algorithm}
\usepackage{algpseudocode}
\usepackage{amsmath}
\usepackage{placeins}
\graphicspath{{media/}}% organize your images and other figures under media/ folder

\title{Enhancing IoT based Plant Health Monitoring through Advanced Human Plant Interaction using Large Language Models and Mobile Applications
}

\author{
  Kriti Agarwal \\
  Vellore Institute of Technology \\
  India\\
   \And
  Samhruth Ananthanarayanan\\
  Vellore Institute of Technology \\
  India\\
  \And
  Srinitish Srinivasan \\
  Vellore Institute of Technology \\
  India\\
  \And
  Abirami S \\
  Vellore Institute of Technology \\
  India\\
}

\begin{document}
\maketitle

\begin{abstract}
This paper presents the development of a novel plant communication application that allows plants to "talk" to humans using real-time sensor data and AI-powered language models. Utilizing soil sensors that track moisture, temperature, and nutrient levels, the system feeds this data into the Gemini API, where it is processed and transformed into natural language insights about the plant's health and "mood." Developed using Flutter, Firebase, and ThingSpeak, the app offers a seamless user experience with real-time interaction capabilities. By fostering human-plant connectivity, this system enhances plant care practices, promotes sustainability, and introduces innovative applications for AI and IoT technologies in both personal and agricultural contexts. The paper explores the technical architecture, system integration, and broader implications of AI-driven plant communication.
\end{abstract}

% keywords can be removed
\keywords{Large Language Models \and Mobile Applications \and Plant Health \and Internet of Things \and Human Plant Interaction \and Gemini}

\section{Introduction}
In recent years, the integration of Artificial Intelligence (AI) and Internet of Things (IoT) technologies has opened up innovative avenues for enhancing human interaction with the natural world. One such innovation is the ability to bridge the communication gap between humans and plants, leveraging the capabilities of sensors and AI-driven models. Plants, although non-verbal, possess vital signals that can be harnessed and interpreted using real-time data from sensors. This paper introduces a novel application that enables plants to “talk” to humans through an AI-powered interface using the Gemini API. The system collects real-time soil data and transforms it into insights that users can interpret as the plant’s “mood” and health status, facilitating a unique form of human-plant communication\cite{tabuenca2021talking}

Soil sensors embedded in the plant’s environment are essential to this system. These sensors monitor critical factors such as moisture, temperature, and pH levels, which are fed to the Gemini API in real-time. Recent research has demonstrated the effectiveness of such sensors in accurately monitoring environmental conditions for plants, enabling users to make informed decisions about care and management\cite{khan2021current}. Our app extends this by utilizing AI-driven insights to provide more intuitive feedback to users. Instead of raw data, the AI translates the plant’s needs into a human-readable format, enabling users to track plant health, understand its “mood,” and interact with it directly through a language-based interface\cite{dwivedi2024analyzing}.

The app was developed using Flutter, Firebase, and ThingSpeak, a combination that ensures the system is scalable, responsive, and capable of real-time data processing. Flutter was selected for its cross-platform functionality, allowing the app to be deployed across multiple devices seamlessly. Firebase offers backend services for data storage and real-time updates, while ThingSpeak handles the collection and transmission of IoT sensor data \cite{juliana2021evecurate}. Together, these technologies create a robust ecosystem, allowing for continuous monitoring and real-time plant-human interaction.

This approach offers a new dimension of connectivity between humans and plants, fostering a deeper relationship with nature through digital means. Prior studies have highlighted the growing interest in human-nature interaction facilitated by AI and IoT technologies. By transforming the way we understand and care for plants, this app aims to promote sustainability and more attentive plant care practices, encouraging users to engage with their environment in a meaningful way. The combination of AI and sensor data interpretation can also be applied to agriculture, enhancing crop management and improving yields by providing personalized care insights for plants\cite{shaikh2022towards}

In this paper, we will explore the development of the app, focusing on its technical architecture and system integration. We will also review the broader implications of utilizing AI and IoT in human-plant communication, discussing the potential benefits and limitations. The insights gained from this study contribute to the growing body of research on AI-driven interactions with the natural world, showcasing how emerging technologies can reshape our relationship with the environment\cite{singh2022systematic}

\section{Literature Survey}
The integration of AI and IoT in plant monitoring systems has been a growing area of research, particularly as sensors have become more accessible and precise. Numerous studies have explored the application of soil sensors for real-time environmental monitoring, providing data on factors such as moisture, temperature, and pH levels. For example, Batte et al. demonstrated the effectiveness of soil sensors in continuously monitoring environmental conditions, highlighting their potential to enhance agricultural practices and home gardening \cite{batte2021}. These sensors have since been integrated into larger IoT systems, where real-time data transmission is essential for immediate feedback, laying the groundwork for interactive applications like the one developed in this project.

Artificial Intelligence has also been increasingly employed to interpret sensor data and provide users with actionable insights. Research by Jain et al. explored the potential of AI models to translate raw environmental data into intuitive, human-readable outputs, thereby enhancing the interaction between humans and plants \cite{jain2020}. This approach is central to the design of our application, which uses the Gemini API to process real-time sensor data and convert it into natural language feedback for the user. Santos and Barros further explored the application of AI for plant health monitoring, identifying both the potential benefits and challenges of using AI to facilitate human-plant interaction \cite{santos2023}. These studies underscore the growing role of AI in making complex biological data accessible and meaningful to everyday users.

Furthermore, the integration of cloud-based platforms like Firebase has proven to be a key factor in the scalability and responsiveness of modern applications. Harley and Chesser’s work on Flutter and Firebase highlights the importance of real-time data handling and cross-platform compatibility in ensuring seamless user experiences \cite{harley2020}. Zhou et al. examined how IoT platforms, such as ThingSpeak, enable continuous data streaming and storage, particularly in agricultural contexts \cite{zhou2022}. These technologies were integral to the design of our app, ensuring real-time interaction between users and plants. This body of research provided critical insights into the architectural decisions behind the system, demonstrating how emerging technologies can be harnessed to build intuitive, scalable solutions for plant care.

\section{App Structure}

The app consists of four main sections, each of which plays a critical role in enabling users to communicate with their plants and track their health: 
\begin{itemize}
    \item The front-end
    \item The back-end storage and data retrieval
    \item The sensors
    \item The chatbot
\end{itemize}

\subsection{Front-End}

The front-end of the app is developed entirely using Flutter, a cross-platform framework that ensures the app functions smoothly across Android and iOS devices. Flutter was chosen for its flexibility, real-time rendering capabilities, and ability to create a highly responsive user interface. The front-end is designed to provide an intuitive and seamless experience, where users can easily interact with the app and access vital information about their plants.

User authentication and authorization are handled through Google Play services, ensuring that each user has a secure and personalized experience. When a user logs in, their account is authenticated using their Google credentials, and a corresponding user entry is maintained in the app’s database. This allows for easy tracking of user-specific data and plant profiles, making sure that each user only has access to their own plants.

The interface provides features for monitoring plant health, displaying real-time data, and viewing historical trends for each plant. Users can also receive notifications and alerts based on real-time sensor data, which inform them of their plant’s current needs. This interface was designed to be user-friendly, ensuring that even those without technical expertise can maintain and nurture their plants effectively.

\subsection{Back-End}

The back-end of the application is built using a combination of real-time and NoSQL databases. The back-end architecture ensures that sensor data is efficiently processed, stored, and retrieved, while also maintaining scalability and performance.

The real-time database is primarily responsible for handling sensor data. Each sensor attached to a plant has a unique ID, and data is continuously uploaded to the real-time database using this sensor ID. The microcontroller nodes of the sensor network transmit data directly to the database, ensuring real-time updates. This real-time database serves as a temporary data store for the sensor readings, which are later fetched by the app for analysis. The use of a real-time database allows for instant updates and ensures that users are always presented with the most current information about their plants.

In addition to the real-time database, a separate NoSQL database is used to manage user and plant data. This NoSQL database was chosen for its scalability, making it possible to handle numerous users and a wide variety of plant species efficiently. It stores key information such as user profiles, plant types, and ideal environmental conditions for different plant species. The separation of real-time sensor data and user/plant data ensures the efficient handling of large volumes of information, making the app scalable for broader use. 

Moreover, the data from the real-time database is processed to determine the plant's mood, which is an essential feature for enhancing user engagement. This mood data is further utilized in the chatbot system, which provides a humanized experience by interpreting sensor readings in a way that users can easily understand.

\subsection{Sensors}

The sensors attached to the plant’s soil play a critical role in gathering real-time data about the plant's environment. These sensors measure parameters such as soil moisture, temperature, humidity, and sunlight exposure, which are vital for determining the plant’s health and well-being. 

Each sensor is connected to a microcontroller node equipped with Wi-Fi capabilities, allowing it to transmit the collected data to the real-time database. The microcontroller processes the raw sensor readings, converting them into digital values or percentages that are readable and interpretable by the app. For instance, soil moisture might be expressed as a percentage of ideal moisture content, while temperature and humidity are displayed in Celsius and percentage, respectively.

This system ensures that the data is not only collected but also standardized, enabling consistent and accurate analysis. These environmental factors are continuously monitored, and the app aggregates this data to present a comprehensive view of the plant’s current health status. The use of multiple sensors ensures that the app can provide precise feedback, and this feedback is essential for both the user and the chatbot interface.

\subsection{Chatbot}

One of the most unique features of the app is the chatbot, which allows users to "talk" to their plants. The chatbot is powered by a large language model (LLM) that leverages the sensor data to create a natural conversation between the user and the plant. The chatbot translates the plant’s health and environmental data into human-like responses, providing the user with real-time updates on how their plant is doing.

The chatbot utilizes prompt engineering techniques to interpret the data collected from the sensors and communicate it in a way that is understandable and engaging. For instance, if the soil moisture is low, the chatbot might say, "I'm feeling a bit thirsty today, could you water me?" Similarly, if the plant is in ideal conditions, the chatbot could respond with, "I'm feeling great, thanks for taking care of me!" This conversational interface humanizes the plant, making it more relatable to the user.

By enabling this form of interaction, the chatbot enhances user engagement and makes the process of plant care more enjoyable. Users can check in with their plants, ask them how they are doing, and receive helpful suggestions for maintaining optimal conditions. This not only makes plant care more interactive but also encourages users to be more attentive to their plants’ needs. Additionally, the chatbot's responses are continuously updated based on real-time data, ensuring that the user is always receiving the most accurate and relevant information.

\section{Conclusion}

In summary, the app's architecture consists of a user-friendly front-end built with Flutter, a robust back-end for data management using real-time and NoSQL databases, sensor networks that continuously monitor environmental factors, and a chatbot that facilitates human-plant interaction. Together, these components work in unison to create a system that helps users maintain their plants' health while fostering a unique, engaging relationship between humans and their plants. The integration of real-time data, AI, and IoT technologies has the potential to transform plant care by making it more interactive, intuitive, and accessible.

\section{Results}

\begin{table}[h!]
    \centering
    \caption{Human Plant Interaction Examples}
    \scalebox{0.65}{\begin{tabular}{|c|c|}
    \hline
         Prompt&Response \\\hline
           Hey how are you doing today?& Greetings I am thriving in this cozy environment. The temperature and humidity are spot on and I am feeling hydrated and energized. Thanks for checking in.\\\hline

           Do you want any water?& I have got plenty of water. I dont think I need any more at the moment. Thanks!\\\hline
    \end{tabular}}
    \label{tab:interaction}
\end{table}
\subsection{Development Environment}
The app was built using Flutter(3.24) using the Android Studio development toolkit. The application was developed on an HP Pavilion Notebook with 16GB RAM and 512 GB internal storage while the app was tested on an OnePlus Nord 3 with 8GB RAM, 128GB internal storage with a resolution of 2272X1240 pixels, 450ppi. The sensors used in all experiments were acquired from an Internet of Things(IoT) laboratory of Vellore Institute of Technology, Chennai.

\subsection{Readings for LLM Interpretation}
\begin{figure}
    \centering
    \includegraphics[width=0.5\linewidth]{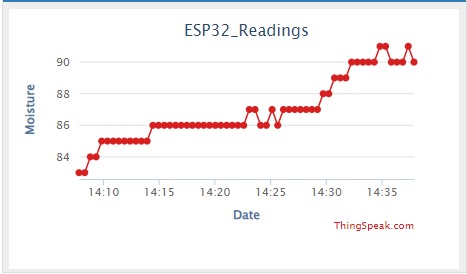}
    \caption{Soil Moisture Readings}
    \label{fig:soil}
\end{figure}

\begin{figure}
    \centering
    \includegraphics[width=0.5\linewidth]{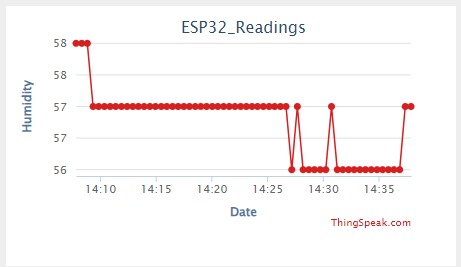}
    \caption{Humidity Readings}
    \label{fig:hum}
\end{figure}

\begin{figure}
    \centering
    \includegraphics[width=0.5\linewidth]{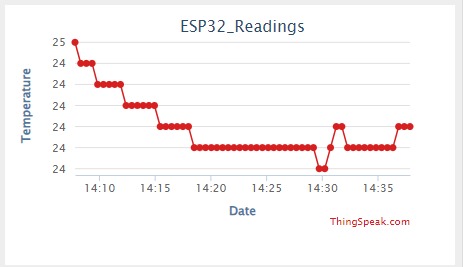}
    \caption{Temperature Readings}
    \label{fig:temp}
\end{figure}

In this section, we present the readings of the sensors. Considering the constraints in equipment at the laboratory, we made use of only soil moisture and DHT11 sensors. Figures \ref{fig:soil} to \ref{fig:temp} detail the readings obtained.

\subsection{Prompt Response Analysis}
For our conditioned experiments, we consider 2 scenarios
\begin{itemize}
    \item Plants on soil with no moisture content
    \item Plants on soil with moisture content
\end{itemize}
The prompts have been engineered to bring in the context of a plant human interaction. For instance, if the plant considered is a Cactus, the LLM is prompted by "Imagine you are a cactus....". Such a dynamic technique ensures that no resources were used on fine-tuning. Table \ref{tab:interaction} lists the prompts and responses for experiments conducted.

%Bibliography
\bibliographystyle{unsrt}
\FloatBarrier
\bibliography{paper}

\end{document}